# Constrained Motion Planning for a Robotic Endoscope Holder based on Hierarchical Quadratic Programming

Jacinto Colan
*Department of Micro-Nano Mechanical Science and Engineering*
*Nagoya University*
Nagoya, Japan
colan@robo.mein.nagoya-u.ac.jp

Ana Davila
*Institutes of Innovation for Future Society*
*Nagoya University*
Nagoya, Japan
davila.ana@robo.mein.nagoya-u.ac.jp

Yasuhisa Hasegawa
*Department of Micro-Nano Mechanical Science and Engineering*
*Nagoya University*
Nagoya, Japan
hasegawa@mein.nagoya-u.ac.jp

*Abstract*—Minimally Invasive Surgeries (MIS) are challenging for surgeons due to the limited field of view and constrained range of motion imposed by narrow access ports. These challenges can be addressed by robot-assisted endoscope systems which provide precise and stabilized positioning, as well as constrained and smooth motion control of the endoscope. In this work, we propose an online hierarchical optimization framework for visual servoing control of the endoscope in MIS. The framework prioritizes maintaining a remote-center-of-motion (RCM) constraint to prevent tissue damage, while a visual tracking task is defined as a secondary task to enable autonomous tracking of visual features of interest. We validated our approach using a 6-DOF Denso VS050 manipulator and achieved optimization solving times under 0.4 ms and maximum RCM deviation of approximately 0.4 mm. Our results demonstrate the effectiveness of the proposed approach in addressing the constrained motion planning challenges of MIS, enabling precise and autonomous endoscope positioning and visual tracking.

## I. Introduction

Minimally Invasive Surgery (MIS), compared to traditional open surgery, reduces scarring and recovery time by accessing the patient's anatomy through small incisions or natural orifices, using an endoscope and long surgical instruments. However, this approach introduces technical challenges for the surgeon, including reduced intuitiveness for precise tool control, limited visibility of tissues and organs, and lack of force feedback [1]. Robot-assisted minimally invasive surgical (RMIS) systems have been shown to be useful in overcoming these limitations with 3d vision, intuitive human-robot interfaces, and high-dexterity surgical tools [2].

Precise endoscope positioning is essential for visualization in the constrained surgical workspace of MIS. Robotic endoscope holders have been proposed to facilitate control of the camera view [3]. Intuitive control of the robotic system can be achieved through physical interaction or remote human-robot interfaces [4]. For example, the da Vinci surgical system, the most widely used robotic surgical system, allows control of endoscope position via a teleoperated robotic arm and user interface. However, alternating between control of surgical tools and the endoscope increases cognitive workload, and therefore some degree of autonomy is desired.

Autonomy in robotic endoscopes has primarily focused on enabling autonomous visual target tracking [5]. Once visual target recognition is achieved, a visual servoing control strategy is commonly employed to generate a trajectory for tracking the target. However, kinematic constraints must be taken into account during motion planning. For instance, in RMIS, the endoscope must adhere to the RCM constraint at the trocar insertion point to prevent tissue damage. This constraint can be enforced either by passive mechanisms or programmable virtual constraints in software [6]. With the increasing use of general-purpose industrial robotic manipulators in surgical applications [7], programmable virtual RCMs offer greater flexibility and versatility for meeting specific surgical task requirements. Therefore, their use is preferred over passive mechanisms.

This work proposes an optimization-based approach for motion planning of a robotic endoscope in minimally invasive surgery. The approach uses Hierarchical Quadratic Programming (HQP) to simultaneously satisfy kinematic constraints and achieve a visual servoing task. A Remote Center of Motion (RCM) constraint is prioritized as a high-priority task to ensure safety, while visual servoing to track a visual target is treated as a lower-priority task.

## II. Related Works

### A. Constrained Motion Planning in RMIS

A common approach to addressing RCM constraints is the hierarchical projection method [8], where RCM constraints are defined as high-priority tasks. For example, Azimian et al. [9] implemented RCM constraints by projecting an end-effector tracking task into the null space of the RCM task. Aghakani et al. [10] proposed a generalized kinematic formulation of the RCM constraint, augmenting the end-effector tracking task Jacobian with the RCM task Jacobian to solve both tasks simultaneously. However, this approach requires defining an endoscope insertion velocity and can lead to algorithmic singularities. Sandoval et al. [11] proposed an RCM implementation for torque-controlled redundant manipulators based on minimum tool distance to the insertion point independent of insertion velocity. Marinho et al. [12] proposed generating a trajectory by projecting a desired endoscope position from the perspective of the RCM, then solving the inverse kinematics problem to compute the corresponding joint configuration. Most previous approaches have followed a null-space projection approach, in which additional constraints, such as joint limits or collision avoidance, cannot easily be incorporated. In contrast, optimization-based methods have been proposed [13, 14, 15] to more easily include additional constraints and avoid local minima, but at the cost of higher computation times.

### B. Multi-objective Hierarchical Optimization

It is common in RMIS to perform multiple tasks simultaneously. The RCM constraint is typically considered the highest-priority task, while visual tracking or tool manipulation are lower-priority tasks that must not violate the RCM constraint. The hierarchical null space projection method [16] is a common approach but can suffer from local minima, singularities, and inability to handle inequality constraints. Hierarchical optimization frameworks can address these challenges. A weighted strategy can define task importance so that the optimal solution depends on the weights, but careful tuning is required and strict prioritization is not possible. In contrast, hierarchical quadratic programming (HQP) solves a quadratic programming problem at each task level and ensures that solutions for lower-priority tasks do not affect higher-priority tasks [17]. HQP has shown promising results in humanoid control, with fast and accurate handling of highly redundant robots [18, 19]. It has also been applied to surgical robots to maximize manipulability for manipulation tasks [20] and constrained inverse kinematics of multi-DOFs surgical tool [21].

### III. CONSTRAINED MOTION PLANNING

We proposed a framework for constrained motion planning for endoscope positioning based on a hierarchical optimization framework.

#### A. Kinematic Formulation of the Remote Center of Motion

We define the RCM constraint as the high-priority task, which keeps the endoscope shaft pivoting around the trocar insertion point. The RCM constraint minimizes the distance between the trocar point $p_{trocar}$ and the closest point on the endoscope shaft $p_{rcm}$, as shown in Fig. 1. The point $p_{rcm}$ is computed as:

$$p_{rcm} = p_{pre} + p_r^T \hat{p}_s \hat{p}_s , \quad (1)$$

where $p_{pre}$ is the position of the joint before the RCM, $p_r$ is the vector from $p_{pre}$ to $p_{trocar}$, and $\hat{p}_s$ denotes the unit vector along the endoscope shaft. The RCM task Jacobian $J_{rcm} \in \mathbb{R}^{1 \times n}$ can be computed as [14]:

$$J_{rcm} = p_e^T \frac{\delta p_{rcm}}{\delta q} \quad (2)$$

$$= p_e^T \left[ (I - \hat{p}_s \hat{p}_s^T) J_{pre} + (\hat{p}_s p_r^T + p_r^T \hat{p}_s I) \frac{\delta \hat{p}_s}{\delta q} \right]$$

with

$$\frac{\delta \hat{p}_s}{\delta q} = \frac{1}{\|p_s\|} (I - \hat{p}_s \hat{p}_s^T)(J_{pre} - J_{post}) \quad (3)$$

where $J_{pre}$ and $J_{post}$ are the analytical Jacobian matrices of the immediate joints before and after the RCM respectively.

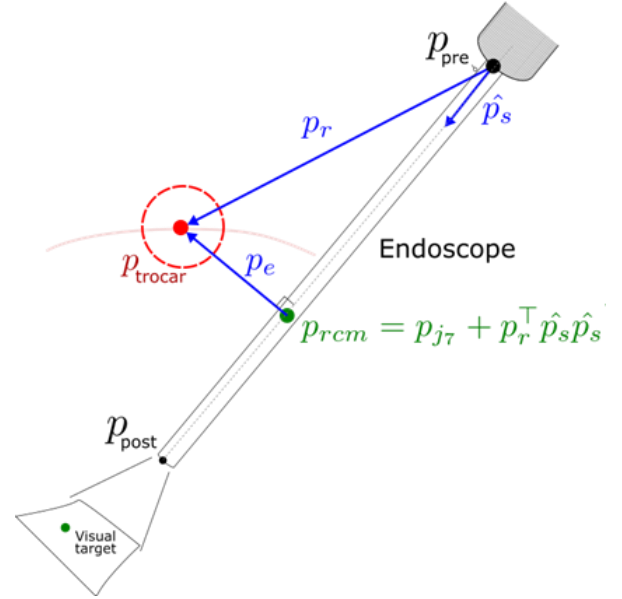

Fig. 1. Remote-center-of motion characterization for a robot-assisted endoscope positioning.

#### B. Hierarchical Quadratic Programming

In general, the inverse kinematics (IK) problem of a manipulator with $n$ degrees of freedom can be represented by:

$$\dot{q} = J_{(q)}^\dagger \dot{x}_{des} \quad (4)$$

where $\dot{q} \in \mathbb{R}^n$ is the desired joint velocity, $\dot{x}_{des} \in \mathbb{R}^m$ represents the desired task space velocity, $J_{(q)} \in \mathbb{R}^{m \times n}$ is the configuration-dependent task Jacobian relating joint velocities with task space velocities, and $J_{(q)}^\dagger$ denotes the pseudoinverse of $J_{(q)}$. Equation (4) is a specific solution of the generalized least square minimization problem:

$$\dot{q}^* = \arg\min_{\dot{q}} \left\| J_{(q)} \dot{q} - \dot{x}_{des} \right\|^2 \quad (5)$$

We are interested in the case in which $m > n$, i.e. the manipulator is redundant with respect to the task. This allows to define additional tasks with lower priorities that can be executed without perturbing the tasks with higher priority. In general, considering $k$ task, each with different level of priority, the multi-objective optimization defined as nested least square problems, each defined as:

$$\min_x \frac{1}{2}\|A_k x - b_k\|^2 \quad (6)$$

$$s.t. \quad C_1 \leq d_1, \ldots, C_k \leq d_k$$
$$E_1 \leq f_1, \ldots, E_k \leq f_k$$

where $A_k, C_k, E_k$ are generic matrices, $b_k$ a generic vector, and $x$ represent the optimization variable subjected to the given constraints. To avoid affecting higher priority tasks, previous $1, \ldots, k-1$ task solutions must keep the optimality conditions defined as $A_{k-1}x = A_k x^*_{k-1}$.

For robot-assisted endoscope positioning, we consider two tasks: RCM constraint with high priority and a visual tracking task with low priority. The corresponding minimization formulation for the RCM constraint problem is defined as follows:

$$\min_{\dot{q}} \|J_{rcm}\dot{q} - e_{rcm}\|^2 \quad (7)$$

$$s.t. \quad \frac{q^- - q}{\delta t} \leq \dot{q} \leq \frac{q^+ - q}{\delta t}$$

with $e_{rcm}$ denotes the minimum distance between the endoscope shaft and the trocar, and can be computed as $e_{rcm} = \|p_{trocar} - p_{rcm}\|$, $q^-$ and $q^+$ are the manipulator joint limits, and $\delta t$ is the time for one control cycle. The use of a slack variable $w = [w^+ \; w^-]$. allows for independent joint limits inequality constraints:

$$\min_{\dot{q}} \|J_{rcm}\dot{q} - \dot{e}_{rcm}\|^2 + \frac{1}{2}\|w\|^2 \quad (8)$$

$$s.t. \quad \dot{q} - \frac{q^+ - q}{\delta t} \leq w^+$$
$$-\dot{q} + \frac{q^- - q}{\delta t} \leq w^-$$

Equation (8) can then be represented as a QP problem [16]:

$$x_1^* = \min_x \frac{1}{2} x^T Q_1 x + p_1^T x \quad (9)$$

$$s.t. \quad [C_1 \; -I]x \leq d_1$$

where $x_1^* = [\dot{q}_1^* \; w^*]$ represents the optimal solution for the minimization problem, $\dot{q}_1^* \in \mathbb{R}^n$ is the optimal joint velocity for the RCM constraint task, $Q_1 \in \mathbb{R}^{3n \times 3n}$, $p_1 \in \mathbb{R}^{3n \times 1}$, $C_1 \in \mathbb{R}^{2n \times n}$, and $d_1 \in \mathbb{R}^{2n \times 1}$ are computed as

$$Q_1 = \overline{A_1}^T \overline{A_1}, \quad (10)$$

$$p_1 = -\overline{A_1}^T \overline{b_1} \quad (11)$$

$$C_1 = \begin{bmatrix} I \\ -I \end{bmatrix} \quad (12)$$

$$d_1 = \begin{bmatrix} \frac{q^+ - q}{\delta t} \\ -\frac{q^- - q}{\delta t} \end{bmatrix} \quad (13)$$

where

$$\overline{A_1} = \begin{bmatrix} J_{rcm} & 0 \\ 0 & I \end{bmatrix} \quad (14)$$

$$\overline{b_1} = \begin{bmatrix} e_{rcm} \\ 0 \end{bmatrix} \quad (15)$$

By solving the QP problem, an optimal solution $\dot{q}_1^*$ is obtained. For the visual tracking task, we need to find an optimal solution $\dot{q}_2^*$ such that it does not affect the solution $\dot{q}_1^*$, this can be achieved by considering the null space of the Jacobian matrix $J_{rcm}$ defined as

$$N_1 = \left(I - J_{rcm}^\dagger J_{rcm}\right). \quad (16)$$

The second QP problem is formulated as follows.

$$\min_{\dot{q},w} \|J_{vis}(N_1\dot{q} + \dot{q}_1^*) - \dot{e}_{vis}\|^2 + \frac{1}{2}\|w\|^2 \quad (17)$$

$$s.t. \quad (N_1\dot{q} + \dot{q}_1^*) - \frac{q^+ - q}{\delta t} \leq w^+$$
$$-(N_1\dot{q} + \dot{q}_1^*) + \frac{q^- - q}{\delta t} \leq w^-$$

with $J_{vis} \in \mathbb{R}^{2 \times n}$ is the Jacobian matrix for the visual tracking task, and $e_{vis}$ the error of the visual feature position in pixels with respect to the center of the endoscope image [22]. Equation (17) is then formulated as a QP problem:

$$x^* = \min_x \frac{1}{2} x^T Q_2 x + p_2^T x \quad (18)$$

$$s.t. \quad [C_2 \; -I]x \leq d_2$$

with $C_2 = C_1$, $d_2 = d_1$, $Q_2 \in \mathbb{R}^{3n \times 3n}$ and $p_2 \in \mathbb{R}^{3n \times 1}$ computed as

$$Q_2 = \overline{A_2}^T \overline{A_2}, \quad (19)$$

$$p_2 = -\overline{A_2}^T \overline{b_2} \quad (20)$$

with

$$\overline{A_2} = \begin{bmatrix} J_{vis} N_1 & 0 \\ 0 & I \end{bmatrix} \quad (21)$$

$$\overline{b_2} = \begin{bmatrix} -(J_{vis}\dot{q}_1^* - e_{vis}) \\ 0 \end{bmatrix} \quad (22)$$

When solving the second QP problem, an optimal solution $\dot{q}_2^*$ is obtained. The optimal solution for both tasks is computed as follows:

$$\dot{q}_{sol}^* = N_1 \dot{q}_2^* + \dot{q}_1^* \quad (23)$$

## IV. EXPERIMENTAL VALIDATION

### A. Robotic System

We validated the proposed approach on a robotic system comprising a 6-DOF robotic manipulator (VS050, Denso Robotics), with a camera (MQ022CG, Ximea) and a rigid endoscope mounted on the end-effector of the manipulator, as shown in Figure 2.

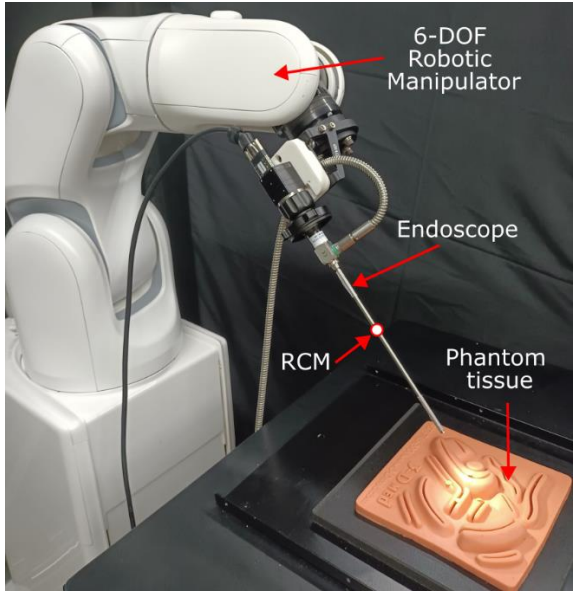

Fig. 2. Robotic endoscope comprising a 6-DOF manipulator holding a camera, and a rigid endoscope. The target markers are placed over a phantom tissue.

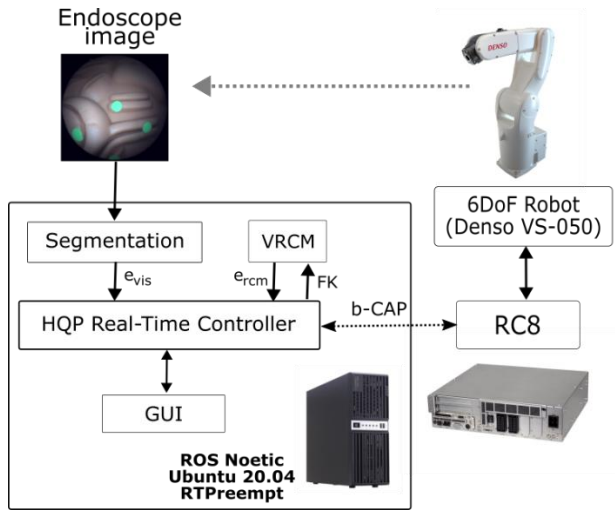

Fig. 3. Overview of the controller architecture.

The control system was implemented on a 2.4 GHz Intel Core i7 computer running Linux (Ubuntu 20.04, Canonical) with real-time patches (RT-PREEMPT) and the Robot Operating System (ROS) framework on top of it. An overview of the controller implemented is depicted in Figure 3. Visual targets are represented as green markers, and are segmented using a color filtering with the OpenCV library. A precalibration of the camera is performed to estimate the intrinsic and extrinsic camera parameters for the visual servoing Jacobian $J_{vis}$. The visual error $e_{vis}$ is defined as the pixel distance between the target marker and image center. The RCM error $e_{rcm}$ is computed at each iteration from joint positions using (1). The HQP controller computes the optimal joint velocity $\dot{q}^*_{sol}$ that satisfies both tasks. The Pinocchio library [23] integrated with the CasADi library [24] was used to generate optimized C code for the kinematics transformations and interface with QP solvers. The OSQP solver [25] solves the QP problem at each task level. The controller communicates with the manipulator at a frequency of 500 Hz through a b-cap protocol. A smooth Cartesian end-effector trajectory was generated online using the Reflexxes Type II motion library [26]. This online trajectory generator (OLT) provides commands for low-level motion control in real time with position and velocity constraints.

### B. Experimental Setup

We evaluated the proposed control framework with a constrained visual tracking task. The manipulator sequentially tracks three visual targets placed over a phantom tissue, starting from target 1. The targets are represented with green circular marker forming a square shape with side length of 20mm as shown in Figure 4A. The initial position of the manipulator is set so that the upper green marker is centered in the endoscope view as shown in Figure 4B. The RCM was set to [0.565 m, 0.0 m, 0.268 m] with respect to the robot base frame. A motion capture (Optitrack) is used to record the shaft position, by placing reflective markers along the endoscope shaft.

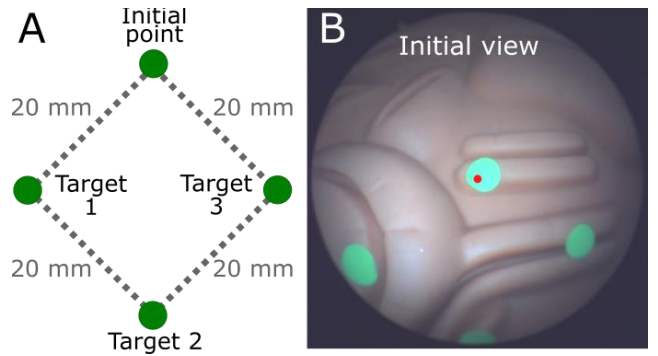

Fig. 4. **A.** Description of the targets position to be tracked. **B.** Initial endoscope view centered over the initial green circle.

### C. Results and Discussion

We evaluated the proposed method in terms of optimization solving time, RCM error, and visual tracking error. The average optimization solving time was 372 μs, demonstrating the real-time capability of the proposed HQP-based approach.

TABLE I. AVERAGE SOLVING TIME

| Time (ms) |
|---|
| 0.372 |

The endoscope shaft path is shown in Figure 5, with the blue lines indicating the shaft and the red circle showing the RCM location. The RCM constraint is maintained as the endoscope moves.

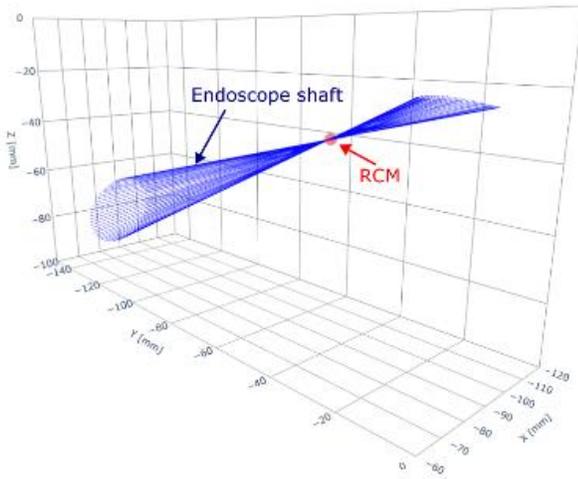

Fig. 5. Endoscope shaft displacement while performing the visual tracking. The RCM location is represented by the red circle.

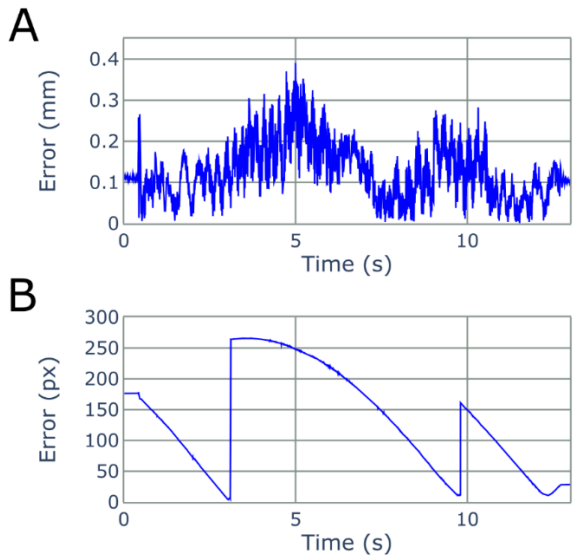

Fig. 6. **A.** RCM error. **B.** Visual error.

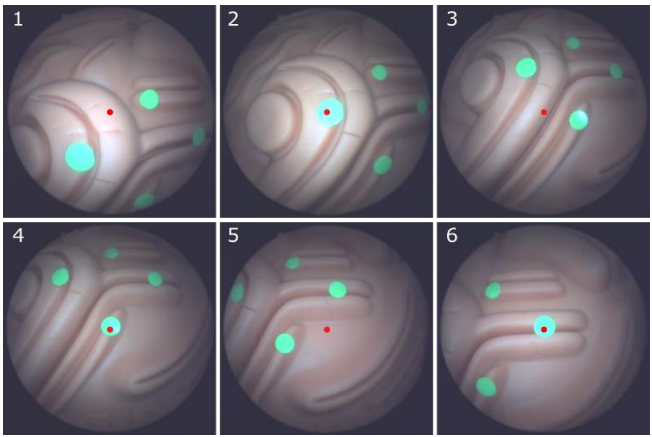

Fig.7. Snapshots of the endoscope view when tracking the visual markers.

Figure 6A shows the time evolution of RCM error, with a maximum of 0.4 mm at t=5s and mean of 0.15 mm, verifying effective RCM constraint preservation. Figure 6B shows the visual error over time, defined as the pixel distance between the marker and image center. Between t=0 and t=3s, the manipulator tracks target 1. Between t=3 s and t=10 s, target 2 is tracked, and after t=10 s, target 3 is tracked. Once the pixel error is under 10 px, the next target is tracked, and the visual error increases accordingly. The visual error decreases roughly linearly for targets 1 and 3, but shows an initial plateau for target 2, likely due to the online trajectory smoother minimizing velocity changes to keep a stable image.

Figure 7 shows endoscope field of view snapshots. Images 1-2 are for target 1, 3-4 for target 2, and 5-6 for target 3, with the image center marked in red. The results demonstrate the precision, safety, and real-time capability of the proposed visual servoing approach under RCM constraints for accurate endoscope control in MIS

## V. CONCLUSION

We presented an optimization-based constrained motion planning scheme for autonomous robotic endoscope control. The proposed hierarchical approach prioritizes an RCM constraint to ensure safety, with a visual tracking task as a lower priority. Both objectives are solved as quadratic program problems online. Experimental results demonstrate real-time capability, high-precision RCM maintenance (maximum deviation 0.4 mm), and tracking accuracy. Future work will focus on incorporating additional constraints such as joint limits and collision avoidance, as well as more complex visual servoing tasks. The framework's flexibility and real-time performance show promise for addressing the technical challenges of autonomous robotic endoscope control in minimally invasive surgery.

### ACKNOWLEDGMENT

This work was supported in part by the Japan Science and Technology Agency (JST) CREST including AIP Challenge Program under Grant JPMJCR20D5, and in part by the Japan Society for the Promotion of Science (JSPS) Grants-in-Aid for Scientific Research (KAKENHI) under Grant 22K14221.